# Adaptive and Detection-Aware Super-Resolution for Small Object Detection

Cezica-Elena LUNCANU and Stelian SPÎNU

***Abstract*—Image super-resolution aims to reconstruct a high-resolution image from a low-resolution input and is relevant for surveillance, satellite imagery, medical imaging, and object detection. This paper investigates SRCNN-based single-image super-resolution for a scale factor of ×4, starting from a classical baseline and progressively introducing residual learning, Charbonnier loss, and adaptive loss functions.**

**The proposed approach includes a three-region adaptive loss, where image areas are divided into smooth, texture, and edge regions using gradient information. The models are evaluated using PSNR, SSIM, LPIPS, and EdgeMAE in order to measure both reconstruction quality and structural detail preservation.**

**In addition to reconstruction evaluation, the paper analyzes the impact of super-resolution on object detection. A YOLOv8 detector is applied to images processed through bicubic interpolation and SRCNN-based variants, using images from the COCO validation dataset. The results show that improved reconstruction quality does not always lead to better detection performance. However, for small objects, a detection-aware variant combined with residual scaling improves the F1 score compared to the bicubic baseline.**

**These results highlight the importance of task-aware evaluation when super-resolution is used in computer vision pipelines.**

## I. Introduction

Digital images are essential in many application domains, including video surveillance, medical imaging, satellite image analysis, driver assistance systems, object recognition, and security-related computer vision tasks [1-2]. In practical scenarios, the available images are often affected by acquisition limitations, poor illumination, compression, motion blur, sensor noise, or other degradation factors. As a result, the captured images may have low resolution, while important visual details become difficult to observe or are completely lost.

This limitation affects both human interpretation and the performance of automatic image analysis algorithms. In object detection, for example, the loss of fine structural details may lead to incorrect classifications, inaccurate localization, or missed detections, especially when the objects are small or located at a significant distance from the camera [3-4].

Image super-resolution aims to reconstruct a high-resolution image from a low-resolution input [5]. Single-image super-resolution is particularly challenging because the missing details must be estimated from only one degraded image. Therefore, super-resolution is commonly treated as an ill-posed inverse problem, since multiple high-resolution images may correspond to the same low-resolution observation.

Classical upscaling methods, such as nearest-neighbor, bilinear, and bicubic interpolation, are simple and computationally efficient. However, these methods cannot recover the real details lost during degradation. Instead, they estimate new pixel values based on local mathematical rules. Although bicubic interpolation remains a widely used baseline, the resulting images may contain blurred edges, weak textures, and limited structural detail [5-6].

Deep neural networks have improved the performance of image super-resolution by learning complex relationships between low-resolution and high-resolution image pairs [5], [7-12]. Convolutional neural networks are particularly suitable for this task because they can extract local visual features such as edges, contours, textures, and object structures. Among these models, SRCNN (Super-Resolution Convolutional Neural Network) [5] represents a simple and interpretable architecture that provides a useful starting point for controlled super-resolution experiments.

This paper investigates how an SRCNN-based architecture can be improved through controlled changes in the training strategy and loss function. The study focuses on ×4 super-resolution and evaluates residual learning, Charbonnier loss, and adaptive loss functions that emphasize visually important regions. In particular, a three-region adaptive loss is analyzed, where image areas are divided into smooth, texture, and edge regions using gradient information.

Beyond reconstruction quality, this paper also examines whether super-resolution improves object detection performance. This aspect is important because, in real computer vision pipelines, super-resolved images are often used as input for automatic analysis systems rather than being evaluated only visually. For this reason, the proposed evaluation includes both classical reconstruction metrics and YOLO-based object detection metrics on COCO validation images [1-2], [13-14].

The main contributions of this paper are summarized as follows:

1. The implementation and evaluation of a complete SRCNN-based pipeline for ×4 image super-resolution.

2. The proposal and evaluation of an adaptive loss function based on three image regions: smooth, texture, and edge.

C.-E. LUNCANU is with the Artificial Intelligence for Defense and National Security master’s program, Faculty of Information Systems and Cyber Security, Military Technical Academy “Ferdinand I”, Bucharest, Romania (e-mail: elena.luncanu@mta.ro).

S. SPÎNU is with the Faculty of Information Systems and Cyber Security, Military Technical Academy “Ferdinand I”, Bucharest, Romania (e-mail: stelian.spinu@mta.ro).

3. The experimental demonstration that improvements in reconstruction metrics, such as PSNR and SSIM, do not automatically imply improved object detection performance.

4. The proposal of a Detection-Aware Super-Resolution variant designed to preserve information relevant to the object detector.

5. The introduction of a Residual Scaling mechanism to control the intensity of the corrections applied to the bicubic image.

6. The experimental demonstration of the benefits of the Detection-Aware + Residual Scaling approach for small objects from the COCO dataset [1].

The implementation discussed in this paper is publicly available on GitHub at the following repository: https://github.com/cezica/adaptive-detection-aware-super-resolution.

## II. Related Work

Image super-resolution has been studied using both classical interpolation methods and deep learning approaches. Traditional upscaling techniques, such as nearest-neighbor, bilinear, and bicubic interpolation, estimate missing pixel values using fixed mathematical rules. These methods are computationally efficient and easy to apply, but they are limited in their ability to recover real high-frequency details lost during the degradation process. For this reason, bicubic interpolation is commonly used as a baseline in super-resolution experiments.

Deep learning methods have significantly improved single-image super-resolution by learning mappings between low-resolution and high-resolution image pairs. SRCNN was one of the first convolutional neural network architectures proposed for this task and demonstrated that a simple end-to-end CNN can outperform classical interpolation methods [5]. The model uses a low-resolution image upsampled through bicubic interpolation and learns to reconstruct a high-resolution output through a sequence of convolutional layers. Due to its simple and interpretable structure, SRCNN remains a useful baseline for controlled experiments and architectural modifications.

Later super-resolution methods introduced deeper architectures, residual learning, perceptual losses, and adversarial training in order to improve visual quality and recover sharper details [7-12]. Models such as EDSR and GAN-based approaches improved perceptual reconstruction quality, but they also increased computational complexity and may introduce artifacts that affect downstream computer vision tasks [9-11]. In this context, lightweight and interpretable SRCNN-based variants remain relevant when the objective is to analyze the effect of specific training strategies, such as residual learning or adaptive loss functions.

Standard datasets such as DIV2K, Set5, and Set14 are commonly used for training, validation, and benchmarking in single-image super-resolution research [6], [15-16].

The choice of loss function is also important in super-resolution. Mean Squared Error is widely used because it is directly related to PSNR, but it may encourage overly smooth reconstructions. L1 loss and Charbonnier loss are often preferred when sharper details and more stable training are desired. In addition, adaptive or spatially weighted losses can give higher importance to visually relevant areas, such as edges and textures. This idea is particularly important because uniform regions, textured areas, and edge regions do not have the same impact on visual quality or on object detection.

Super-resolution is commonly evaluated using reconstruction metrics such as PSNR and SSIM, while perceptual similarity can be measured using LPIPS [17-18]. However, these metrics do not always indicate whether the reconstructed image is useful for a downstream task. In object detection pipelines, a visually improved image may still reduce detection performance if it introduces artifacts or changes local structures used by the detector. Therefore, recent task-aware approaches evaluate super-resolution not only through reconstruction quality, but also through its impact on object detection metrics such as precision, recall, F1 score, and AP.

This paper follows this task-aware perspective by evaluating SRCNN-based super-resolution models both as image reconstruction methods and as preprocessing steps for YOLOv8 object detection. The analysis focuses especially on small objects from the COCO validation dataset, where low resolution can remove important visual details and make detection more difficult.

## III. Methods

This section describes the proposed SRCNN-based super-resolution pipeline and the evaluation procedure used to analyze its impact on object detection. The methodology follows a progressive structure: first, a baseline SRCNN model is implemented for $\times 4$ super-resolution; then, adaptive loss functions are introduced to improve the reconstruction of relevant image regions; finally, the super-resolved images are evaluated as preprocessing inputs for YOLOv8 object detection.

### A. Data Preparation and Degradation Model

The super-resolution models are trained using high-resolution images. For each input image, a low-resolution version is generated artificially by bicubic downsampling with a scale factor of $\times 4$. The low-resolution image is then upsampled back to the original spatial size using bicubic interpolation. This bicubic-upsampled image is used as input for the SRCNN-based models, while the original high-resolution image is used as the target.

The training process is applied on the luminance channel. Each RGB image is converted to the YCbCr color space, and the model is trained only on the Y channel. This approach follows the common practice used in SRCNN-based super-resolution, where most structural information is contained in the luminance component [5]. During inference, the reconstructed Y channel is merged back with the original chrominance channels in order to obtain the final RGB image.

For the reconstruction experiments, the models are evaluated using standard super-resolution metrics [17-18]. For the detection experiments, images from the COCO 2017 validation dataset [1] are used together with their ground-truth annotations.

### B. SRCNN-Based Super-Resolution Model

The baseline model follows the classical SRCNN design, using three convolutional layers. The first layer extracts local image features using a large convolutional kernel, the second layer performs nonlinear feature mapping, and the final layer reconstructs the output luminance image.

In the final ×4 configuration, the SRCNN-style model uses three convolutional layers with 128 and 64 intermediate feature maps. The general structure is:

1. feature extraction layer: convolution with 9×9 kernels;
2. nonlinear mapping layer: convolution with 1×1 kernels;
3. reconstruction layer: convolution with 5×5 kernels.

The model receives the bicubic-upsampled luminance image as input and predicts either the reconstructed image directly or a residual correction. In the residual configuration, the final super-resolved image is obtained by adding the predicted residual to the bicubic image. This strategy allows the network to focus on learning the missing high-frequency details instead of reconstructing the full image from scratch.

### C. Adaptive Three-Region Loss

A key part of the proposed method is the use of an adaptive loss function based on image regions. Instead of applying the same loss uniformly over the entire image, the proposed approach separates the image into three types of regions:

1. smooth regions;
2. texture regions;
3. edge regions.

The separation is performed using gradient information computed on the high-resolution target image. Pixels with low gradient values are assigned to smooth regions, pixels with medium gradient values are assigned to texture regions, and pixels with high gradient values are assigned to edge regions. The best-performing threshold configuration used in the experiments is denoted as Thr1035, corresponding to gradient thresholds 0.10 and 0.35.

The adaptive loss combines different reconstruction objectives depending on the region type. Smooth areas are easier to reconstruct and are treated with a lower structural priority, while texture and edge areas receive stronger emphasis because they contain high-frequency information that is important for visual quality and object detection. The general form of the loss is:

$$L_{total} = \lambda_s L_{smooth} + \lambda_t L_{texture} + \lambda_e L_{edge},$$

where $L_{smooth}$, $L_{texture}$, and $L_{edge}$ represent the reconstruction losses computed on the corresponding masks. This design encourages the model to preserve structural details without over-penalizing uniform areas.

### D. Detection-Aware Super-Resolution

After evaluating the reconstruction-oriented models, a detection-aware variant is introduced. The motivation for this step is that better reconstruction metrics do not necessarily imply better object detection performance. A super-resolution model may improve PSNR or SSIM while still modifying local structures in a way that negatively affects the detector.

The Detection-Aware Super-Resolution variant is designed to preserve information that is useful for object detection, especially in regions containing small objects and structural contours. This variant starts from the adaptive 3Region configuration and introduces an additional detection-oriented objective. The purpose is not only to minimize the reconstruction error, but also to preserve visual structures that are relevant for the downstream YOLOv8 detector.

The detection-aware model is evaluated against bicubic interpolation and the reconstruction-oriented SRCNN variants. This comparison makes it possible to analyze whether super-resolution acts as a useful preprocessing stage for object detection or whether it introduces artifacts that reduce detection performance.

### E. Residual Scaling

The first detection-aware experiments showed that applying the full SRCNN residual may be too aggressive for object detection. Even when the image appears visually improved, strong residual corrections can introduce artifacts or local distortions that affect the detector.

To address this problem, a residual scaling mechanism is introduced. The final reconstructed image is computed as:

$$SR = Bicubic + alpha * Residual,$$

where *alpha* controls the intensity of the correction applied over the bicubic image. A smaller *alpha* preserves more of the bicubic baseline and reduces the risk of artifacts, while a larger *alpha* applies stronger super-resolution corrections.

Several *alpha* values are tested in the experiments, including 0.05, 0.10, 0.15, and 0.30. The best result is obtained for $alpha = 0.15$, which provides a balance between preserving the bicubic structure and adding useful high-frequency corrections.

### F. YOLO-Based Object Detection Evaluation

To evaluate the practical usefulness of the super-resolution models, a YOLOv8 detector [13-14] is applied to images generated by different preprocessing variants. The evaluated variants are:

1. HR image;
2. bicubic interpolation;
3. 3Region Thr1035 with residual scaling;
4. Detection-Aware Thr1035 with residual scaling.

The detection evaluation uses images from the COCO validation dataset and the corresponding ground-truth annotations. Since the main objective is to analyze the effect of super-resolution on difficult detection cases, the final evaluation focuses on small objects. An object is considered small according to the COCO area criterion, where the object area is lower than 32 × 32 pixels [1].

For each image, the detector predictions are matched with the ground-truth boxes using an Intersection over Union threshold of 0.50, following the standard object detection evaluation setting [1-2]. A prediction is considered correct if it has the same class as the ground-truth object and reaches the IoU threshold. Based on the matching results, precision, recall, and F1 score are computed for each variant.

This evaluation allows the paper to compare reconstruction-oriented performance with task-oriented detection performance and to determine whether the proposed super-resolution variants improve the downstream object detection pipeline.

## IV. Results

This section presents the experimental results obtained for both image reconstruction and object detection. The objective is to evaluate not only whether the proposed SRCNN-based variants improve classical super-resolution metrics, but also whether these improvements are reflected in a downstream object detection task.

### A. *Super-Resolution Reconstruction Results*

The first experiment evaluates the reconstruction quality of the SRCNN-based models. The compared variants include bicubic interpolation, the SRCNN V4 baseline, a weighted loss variant, and the proposed three-region adaptive loss configuration denoted as 3Region Thr1035. The models are evaluated using PSNR, SSIM, LPIPS, and EdgeMAE. The reconstruction results are reported in Table I.

Table I. Super-Resolution Reconstruction Results

| *Model* | *PSNR* | *SSIM* | *EdgeMAE* | *LPIPS* |
|---|---|---|---|---|
| Bicubic | 25.381 | 0.9571 | 0.0915 | 0.4125 |
| SRCNN V4 | 26.017 | 0.9635 | 0.0834 | **0.3290** |
| Weighted Loss | **26.045** | 0.9638 | 0.0827 | **0.3290** |
| 3Region Thr1035 | 26.037 | **0.9643** | **0.0805** | 0.3417 |

As shown in Table I, all SRCNN-based variants improve the reconstruction quality compared to bicubic interpolation. The SRCNN V4 baseline already increases PSNR and SSIM, while the adaptive variants provide additional improvements in structural regions. The 3Region Thr1035 model obtains the best SSIM value and the lowest EdgeMAE value, which indicates better preservation of edge-related information. This behavior is consistent with the purpose of the adaptive loss, which assigns higher importance to texture and edge regions.

Fig. 1 illustrates a qualitative comparison between the evaluated reconstruction variants. The visual results show that the SRCNN-based models reduce the blur introduced by bicubic interpolation and recover sharper local structures. However, the differences between the reconstruction-oriented variants remain moderate, which motivates the additional evaluation from the perspective of object detection.

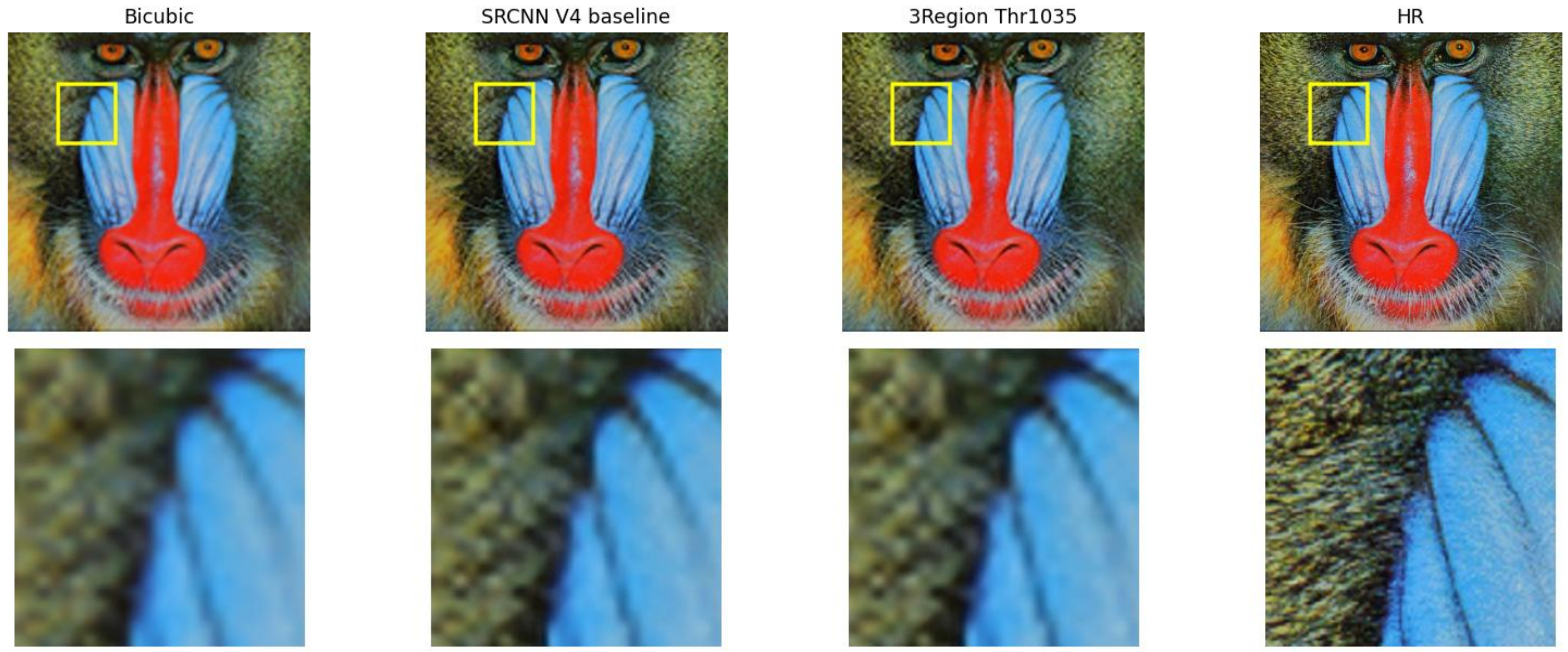

Figure 1. Visual comparison of super-resolution results for bicubic interpolation, SRCNN V4 baseline, 3Region Thr1035, and HR reference

### B. *General Object Detection Evaluation*

The second experiment evaluates whether the reconstruction improvements also improve YOLO-based object detection. The detector is applied to images processed with bicubic interpolation and the SRCNN-based variants. The results are reported using mAP50, AP50, precision, recall, and F1 score. The general detection results are shown in Table II. Although the SRCNN-based models improve reconstruction metrics, they do not outperform bicubic interpolation in the general detection evaluation. Bicubic interpolation obtains the highest mAP50 and F1 score among the compared low-resolution variants. This result confirms that better visual reconstruction does not automatically translate into better object detection performance.

Table II. General YOLO Detection Results

| *Model* | *mAP50* | *AP50* | *Precision* | *Recall* | *F1* |
|---|---|---|---|---|---|
| Bicubic | 0.3768 | 0.3753 | 0.7042 | 0.4231 | 0.5286 |
| SRCNN ×4 Adaptive | 0.3587 | 0.3687 | 0.6952 | 0.4118 | 0.5173 |
| SRCNN ×4 V4 | 0.3556 | 0.3715 | 0.7043 | 0.4133 | 0.5209 |
| 3Region Thr1035 | 0.3421 | 0.3669 | 0.7012 | 0.4104 | 0.5178 |

One explanation is that object detectors are sensitive to local structures, edges, and textures learned during training. A super-resolution model can introduce visually plausible corrections that improve PSNR or SSIM, but these corrections may also create artifacts or modify detector-relevant patterns. Therefore, super-resolution models used in object detection pipelines should be evaluated with task-specific metrics, not only with reconstruction metrics.

### C. *Detection-Aware Super-Resolution and Residual Scaling*

To reduce the negative effect of aggressive super-resolution corrections, residual scaling is introduced. Instead of applying the full residual predicted by the SRCNN model, the final image is computed as the bicubic image plus a scaled residual correction. Several values of the scaling coefficient alpha are tested, and the results are presented in Table III.

TABLE III. EFFECT OF RESIDUAL SCALING ON SMALL-OBJECT DETECTION

| *Alpha* | *F1* |
|---|---|
| 0.05 | 0.4297 |
| 0.10 | 0.4323 |
| 0.15 | **0.4355** |
| 0.30 | 0.4306 |

As shown in Table III, the best F1 score is obtained for $alpha = 0.15$. This value provides the best balance between preserving the bicubic structure and adding useful high-frequency information. A very small alpha applies only a limited correction and therefore has a reduced effect, while a larger alpha may introduce artifacts that affect the detector.

### D. *Final Evaluation on Small COCO Objects*

The final experiment focuses on small objects from the COCO validation dataset. The final small-object detection results are shown in Table IV. This evaluation is relevant because small objects are the most affected by resolution degradation and are more likely to benefit from controlled super-resolution corrections.

As shown in Table IV, the Detection-Aware Thr1035 model with residual scaling obtains the best F1 score among the low-resolution reconstruction variants. Compared to bicubic interpolation, the F1 score increases from 0.4269 to 0.4355. The improvement is moderate, but it shows that task-aware super-resolution combined with controlled residual correction can provide benefits for small-object detection.

TABLE IV. FINAL SMALL-OBJECT DETECTION RESULTS

| *Variant* | *Precision* | *Recall* | *F1* |
|---|---|---|---|
| HR | 0.4178 | 0.5619 | 0.4792 |
| Bicubic | 0.4630 | 0.3961 | 0.4269 |
| 3Region Thr1035 +*alpha* = 0.15 | 0.4619 | 0.4028 | 0.4303 |
| Detection-Aware Thr1035 + *alpha* = 0.15 | **0.4675** | **0.4077** | **0.4355** |

The 3Region Thr1035 model with residual scaling also improves the F1 score compared to bicubic interpolation, but the detection-aware variant obtains better precision, recall, and F1. This indicates that preserving detector-relevant structures is more useful for small-object detection than optimizing only reconstruction quality.

Fig. 2 presents the influence of the residual scaling coefficient alpha on the F1 score. The curve confirms that $alpha = 0.15$ is the best value among the tested configurations.

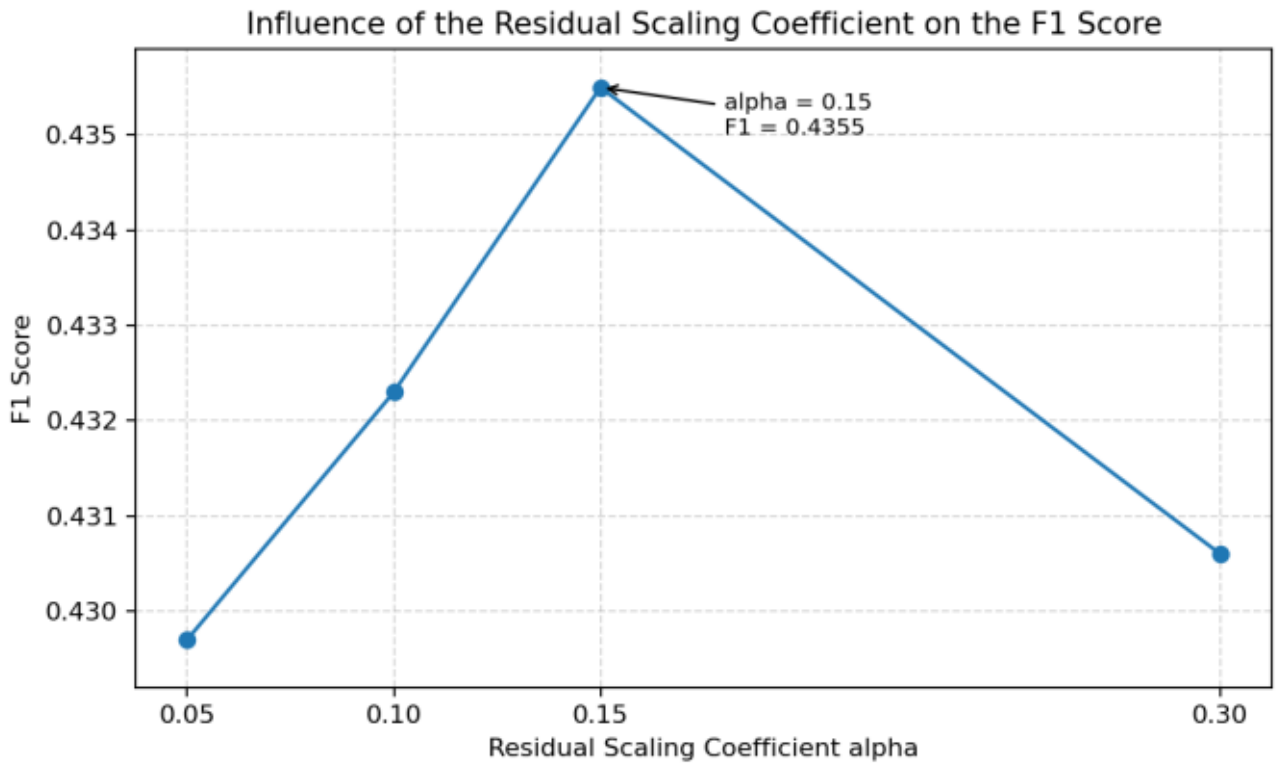

Figure 2. Influence of the residual scaling coefficient alpha on the F1 score for small-object detection

Fig. 3 shows that the Detection-Aware variant increases the number of true positives compared to bicubic interpolation.

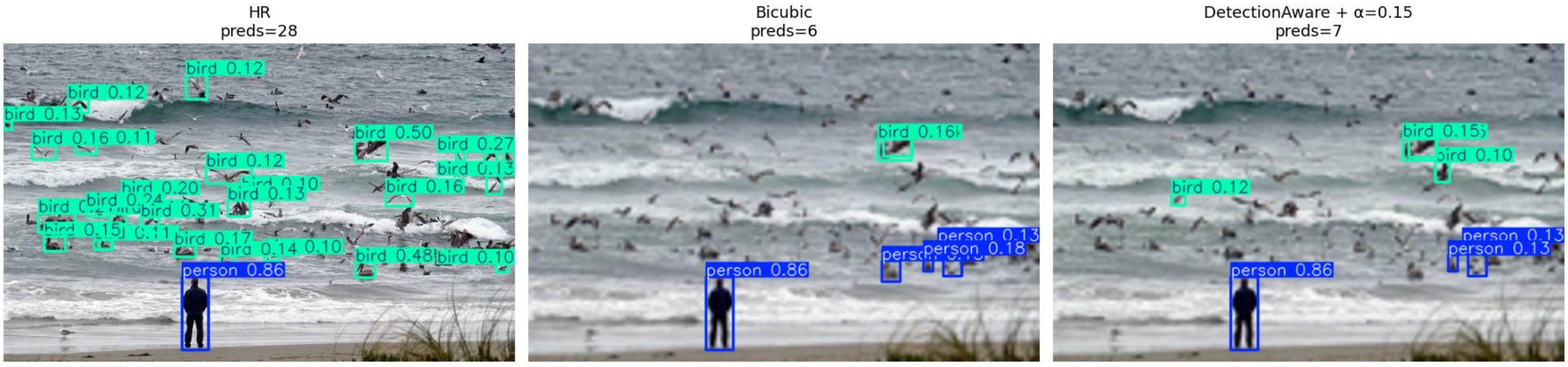

Figure 3. Visual YOLO detection comparison between bicubic interpolation and the Detection-Aware Thr1035 model with residual scaling

### E. *Discussion*

The experiments lead to two main observations. First, adaptive reconstruction losses can improve classical super-resolution metrics and structural detail preservation. Second, these reconstruction improvements do not automatically guarantee better object detection performance. The general YOLO evaluation shows that bicubic interpolation can remain stronger than reconstruction-oriented SRCNN variants, despite lower reconstruction quality.

The final small-object experiment shows that a detection-aware approach combined with residual scaling can improve the downstream detection result. This suggests that super-resolution should be optimized and evaluated according to the final task where it is used. In practical computer vision systems, especially those involving object detection, task-aware metrics are necessary in addition to PSNR, SSIM, and perceptual reconstruction measures.

## V. Conclusion

This paper investigated adaptive and detection-aware super-resolution for small-object detection. Starting from an SRCNN-based architecture, several ×4 super-resolution variants were implemented and evaluated, including residual learning, Charbonnier loss, and an adaptive three-region loss based on smooth, texture, and edge areas.

The reconstruction results showed that SRCNN-based variants improve classical metrics compared to bicubic interpolation. In particular, the 3Region Thr1035 model obtained better structural reconstruction in edge-related regions, as indicated by SSIM and EdgeMAE.

However, the YOLO-based evaluation showed that better reconstruction quality does not automatically lead to better object detection performance. Although the adaptive SRCNN variants improved PSNR and SSIM, they did not consistently outperform bicubic interpolation in the general detection evaluation.

To address this limitation, a Detection-Aware Super-Resolution variant combined with Residual Scaling was evaluated on small COCO objects. The best result was obtained for $alpha = 0.15$, where the Detection-Aware variant improved the F1 score compared to the bicubic baseline.

Overall, the results show that task-aware evaluation is essential when super-resolution is used as a preprocessing step for object detection. Future work may include end-to-end training with the detector, larger COCO evaluations, and testing with more advanced super-resolution architectures.